\documentclass[twoside]{article}

\usepackage[accepted]{aistats2021}
\usepackage[colorlinks=true,linkcolor=blue]{hyperref}%
\usepackage{wrapfig}
\usepackage{url}
\usepackage{colortbl}
\usepackage{amsmath}
\usepackage{soul}
\usepackage{graphicx}
\usepackage{subcaption}
\usepackage{pdfpages}
\usepackage{amsfonts}
\usepackage{tabularx}
\usepackage{booktabs}
\usepackage{algpseudocode}
\usepackage{algorithm}
\usepackage{enumitem}
\usepackage[T1]{fontenc}

\newcommand{\beq}{\vspace{0mm}\begin{equation}}
\newcommand{\eeq}{\vspace{0mm}\end{equation}}
\newcommand{\beqs}{\vspace{0mm}\begin{eqnarray}}
\newcommand{\eeqs}{\vspace{0mm}\end{eqnarray}}
\newcommand{\barr}{\begin{array}}
\newcommand{\earr}{\end{array}}

\newcommand{\Xmat}[0]{{{\bf X}}}

\newcommand{\Zmat}{{\bf Z}}

\newcommand{\hv}[0]{{\boldsymbol{h}} }

\newcommand{\zv}{\boldsymbol{z}}

\newcommand{\Psimat}{\boldsymbol{\Psi}}

\newcommand{\epsilonv}{\boldsymbol{\epsilon}}

\newcommand{\muv}[0]{{\boldsymbol{\mu}}}

\newcommand{\sigmav}[0]{{\boldsymbol{\sigma}} }

\newcommand{\phiv}{\boldsymbol{\phi}}

\newcommand{\given}{\,|\,}

\newcommand*{\emails}[2][@gmail.com]{%
    \def\@tempa{\@gobble}%
    \@for\qrr@email:=#2\do{%
        \edef\@tempb{\noexpand\href{mailto:\qrr@email #1}{\qrr@email}}%
        \edef\@tempa{\unexpanded\expandafter{\@tempa}{, }\unexpanded\expandafter{\@tempb}}}%
    \{\@tempa\}#1%
}

\usepackage{natbib}

\newcommand{\poincare}{Poincar\'{e} }

\begin{document}

%
\runningtitle{Hyperbolic Graph Embedding with Enhanced Semi-Implicit Variational Inference}

%

\twocolumn[

\aistatstitle{Hyperbolic Graph Embedding with Enhanced Semi-Implicit Variational Inference}

\aistatsauthor{ Ali Lotfi Rezaabad \And Rahi Kalantari \And  Sriram Vishwanath \And Mingyuan Zhou \AND Jonathan I Tamir}
\vspace{0.5 cm}
\aistatsaddress{ The University of Texas at Austin}
\vspace{-0.7 cm}
\aistatsaddress{corresponding authors: {\texttt{\{alotfi, jtamir\}@utexas.edu}}}
]

\begin{abstract}
Efficient modeling of relational data arising in physical, social, and information sciences is challenging due to complicated dependencies within the data. In this work we build off of semi-implicit graph variational auto-encoders to capture higher order statistics in a low-dimensional graph latent representation. We incorporate hyperbolic geometry in the latent space through a \poincare embedding to efficiently represent graphs exhibiting hierarchical structure. To address the naive posterior latent distribution assumptions in classical variational inference, we use semi-implicit hierarchical variational Bayes to implicitly capture posteriors of given graph data, which may exhibit heavy tails, multiple modes, skewness, and highly correlated latent structures. We show that the existing semi-implicit variational inference objective provably reduces information in the observed graph. Based on this observation, we estimate and add an additional mutual information term to the semi-implicit variational inference learning objective to capture rich correlations arising between the input and latent spaces. We show that the inclusion of this regularization term in conjunction with the \poincare embedding boosts the quality of learned high-level representations and enables more flexible and faithful graphical modeling. We experimentally demonstrate that our approach outperforms existing graph variational auto-encoders both in Euclidean and in hyperbolic spaces for edge link prediction and node classification.

\end{abstract}

\section{Introduction and Related Work}
Rich relational data represented as undirected graphs arise in a variety of applications spanning physical, social, and information sciences. These include knowledge graphs, gene expression networks, social graphs in social media graphs, citation graphs, recommendation systems, transportation networks, and cellular networks, to name a few. Downstream analyses of these graph data including edge link prediction, clustering, classification, filtering, and denoising are of particular interest and may reveal insight into the underlying processes governing the data. As the number of nodes and edges grows, it is both expensive and challenging to elucidate this information directly from the observations due to computation and complicated structure.

Low-dimensional graph representations aim to address these difficulties by learning a small number of summary statistics of the graph's features, e.g., through unsupervised learning, and using this representation learning for downstream tasks (\cite{bengio}). Initial work in this space focused on deterministic graph embedding in the latent space; see \cite{caiIEEE} for a survey on these methods. More recently, representations have been extended to model uncertainty in the graph by assigning each node to a random variable in latent space (\cite{bojchevski2017deep,shizhu,dossantos,kipf2016variational,svae18,sigvae}). A central idea in many of these works is the use of a tractable approach to estimating the latent posterior distribution through a variational inference framework introduced by \cite{kingma2013auto}. To ease restrictions on the expressibility of the posterior distribution, semi-implicit variational inference (SIVI, \cite{sivi}) and its extension to graph representation (\cite{sigvae}) have been developed to model highly correlated latent structures, thus lifting the need for an explicit probability density function required by classical variational inference techniques.

There remain two additional challenges in succinctly and accurately representing relational data, which we address in this work. First, while SIVI can characterize data exhibiting heavy tails, multiple modes, skewness, and highly correlated latent structures, we show that the existing SIVI objective provably reduces information in the observed graph. This is because the log-likelihood maximization has the undesirable effect of reducing the mutual information between the input observations and latent representation. Thus, while the reconstruction error may be low, the quality of the representation may also suffer. Based on this observation, we estimate and add an additional mutual information term to the SIVI objective to capture the correlation between the input and latent spaces. By optimizing over this term, we are able to capture a more faithful latent representation of the graph.

A second limitation is the use of a Euclidean embedding, which is often not suitable for graphical data exhibiting tree-like hierarchical structure (\cite{adcock,ravasz}), spherical structure (\cite{fisher}), or other complicated dependencies between nodes (\cite{steyvers}. Recognizing this limitation, the authors in \cite{svae18} replace the Gaussian prior with a Mises-Fisher distribution, which imposes a Gaussian distribution on the hypersphere, and thus allows for a hyperspherical embedding. However, as the approach still relies on graph variational auto-encoders (GVAE) consisting of an analytical prior, neither higher-order statistics (e.g. skewness, multi-modality) nor correlations between input and latent space are captured. A number of works have recently explored the utility of hyperbolic latent embeddings for graphs. \cite{nickel2017poincare} show that a deterministic \poincare embedding improved link prediction compared to a Euclidean embedding. Similarly, \cite{poincarevae} extended VAEs to \poincare embeddings and showed promising preliminary results on link prediction compared to the Euclidean counterpart. The work by \cite{poincare_wasserstein} formulated an alternative objective function based on Wasserstein loss in hyperbolic space. Both \cite{nickel2017poincare} and \cite{poincare_wasserstein} considered a broad range of hierarchical data, but did not fully explore downstream applications on graphs. In this work we carry this direction forward and endow our latent space with a hyperbolic \poincare embedding, while also including our enhanced-SIVI framework.

Combining these ideas, we arrive at our \emph{enhanced semi-implicit hyperbolic graph embedding} (ESI-HGE). The main contributions of our work are the following:
\begin{enumerate}[nosep]
    \item We show that the SIVI objective introduced in \cite{sivi} provably reduces the mutual information between the input and latent representation. 
    \item We propose to correct the objective function with an additional mutual information term and provide a computationally tractable approach for estimating it based on Monte Carlo sampling.
    \item We demonstrate empirically that the inclusion of the enhanced SIVI objective in conjunction with a hyperbolic embedding in latent space is able to capture higher-level feature representation and improve quantitative metrics on downstream tasks.
\end{enumerate}

To validate our approach, we compare ESI-HGE to other leading methods for edge link prediction and unsupervised node classification on three popular citation network datasets (\cite{sen:aimag08}). We show that in both tasks, our proposed method outperforms other state-of-the-art methods. We empirically demonstrate that our approach captures more mutual information between input and latent spaces. We also explore the representation quality on synthetically generated hierarchical image-graph data and show that the proposed embedding naturally represents the graph structure. Reference code to reproduce our results is available at \url{https://github.com/utcsilab/ESI_HGE}.

\section{Background}\label{sec:background}
\subsection{Graph VAE}
Introduced by \cite{kipf2016variational}, the graph variational autoencoder (GVAE) is a framework on top of VAEs (\cite{kingma2013auto}) that is capable of embedding attributes of nodes in an undirected graph on to a lower dimensional space. Formally, given an undirected graph $\mathcal{G} = (\mathcal{V}, \mathcal{E})$ with total number of nodes $N = |\mathcal{V}|$, we define $\mathbf{A}$ as its associated adjacency matrix and $\Xmat \in \mathbb{R}^{N\times M}$ as its feature matrix ($M$ features per node). We introduce a stochastic latent variable (of lower dimension) $F$ associated with each node, summarized by the matrix $\Zmat \in \mathbb R^{N\times F}$. Since optimizing the true posterior, $p_{\theta}(\Zmat|\Xmat, \mathbf{A})$, is fairly challenging, GVAEs instead employ a variational posterior $q_{\phi}(\Zmat|\Xmat,\mathbf{A})$, which is parameterized by a graph convolutional network (GCN) with parameters $\phi$:

\begin{align}\label{eq: posterior for GVAE}
\begin{split}
    q_{\phi}(\Zmat|\Xmat, \mathbf{A}) &= \prod_{i=1}^{N} q_{\phi}(\zv_i|\Xmat,\mathbf{A}),\\
    q_{\phi}(\zv_i|\Xmat, \mathbf{A}) &= \mathcal{N}(\zv_i| \boldsymbol{\mu}_i, \text{diag}(\boldsymbol{\sigma}^2_i)), \\
    \boldsymbol{\mu}_i = \text{GCN}_{\mu}(\Xmat, &\mathbf{A}), \quad 
    \log \boldsymbol{\sigma}_i = \text{GCN}_{\sigma}(\Xmat, \mathbf{A}).\\
\end{split}
\end{align}
Here, $\boldsymbol{\mu}_i$ and $\boldsymbol{\sigma}_i$ denote the mean and standard deviation associated with the $i$-th node, respectively. The generative model is given by an inner product of the corresponding higher level representations as follows:

\begin{align}\label{eq: generative for GVAE}
    \begin{split}
        p(\mathbf{A}|\Zmat) = \prod_{i = 1}^{N}\prod_{j=1}^{N} p(A_{i,j}|\zv_i, \zv_j),\\
        p(A_{i,j} = 1|\zv_i, \zv_j ) = \text{sigmoid}(\zv_i^\top \zv_j).
    \end{split}
\end{align}

The optimal values of the GCN parameters $\phi$ can be obtained by maximizing the evidence lower bound (ELBO, \cite{wainwright}):

\begin{equation}\label{eq: ELBO for GVAE}
    \max_{\phi} ~ \mathbb{E}_{q_{\phi}(\Zmat|\Xmat, \mathbf{A})}[\log p(\mathbf{A}|\Zmat)] - \text{KL}(q_{\phi}(\Zmat|\Xmat, \mathbf{A}) ||p(\Zmat)),
\end{equation}
where $\text{KL}$ denotes the Kullback-Leibler (KL) divergence. Furthermore, $p(\Zmat)$ is defined to be a Gaussian prior, i.e.,  $p(\Zmat) = \prod_{i}\mathcal{N}(\zv_i | 0, \mathbf{I})$.

\subsection{Semi-Implicit Variational Inference}
As demonstrated by \cite{sivi}, semi-implicit variational inference (SIVI) provides a computationally simple yet powerful approach to capturing multi-modality and skewness of posterior distributions, while vanilla variational inference is often insufficient. The authors empirically show that SIVI can closely match the accuracy of MCMC, however, with far lower computational cost.  
The latent posterior distribution expresses how a node in the graph is represented in the latent space given its features and adjacency information. In many works (e.g., \cite{bojchevski2017deep,svae18,dossantos,shizhu,kipf2016variational,poincarevae}), the latent representation assumes a uni-modal distribution with uncorrelated latent variables. To enrich this simplistic representation, SIVI develops a posterior representation which captures multi-modality, correlation and skewness in the latent representation of each node. As this representation is more faithful to the data, it improves downstream tasks including edge link prediction and classification using the low dimensional graph representation (\cite{sigvae}).

The semi-implicit VAE enriches the posterior distribution by using a hierarchical generative model where the distribution $h_{\phi}(\Zmat|\Xmat)$ is represented by a mixture model consisting of an infinite number of mixtures with weights given by an implicit distribution $q_{\phi}(\Psimat|\Xmat)$. As a result, the latent variable distribution ($\hv_{\phi}(\Zmat)$) can be written as $\hv_{\phi}(\Zmat|\Xmat)=  \int{q(\Zmat|\Psimat, \Xmat)q_{\theta}(\Psimat|\Xmat) d\Psimat}$. In the semi-implicit VAE,
$\Psimat$ is drawn from an implicit distribution ($q_{\phi}(\Psimat)$) and used as parameter for the exponential family distribution $q_{\phi}(\Zmat|\Psimat, \Xmat)$. To sample from $q_{\phi}(\Psimat \vert \Xmat)$, one generates samples from a simple distribution $q(\epsilonv)$ and maps $\Xmat$ and $\epsilonv$ to $\Psimat$ using $\Psimat = g_{\phi}(\Xmat, \epsilonv)$. Since $\hv_{\phi}(\Zmat|\Xmat)$ is an implicit distribution, it is not possible to explicitly evaluate the ELBO for $\hv_{\phi}(\Zmat|\Xmat)$. Instead, a lower bound ($\underline{\mathcal{L}}_{K}$) is evaluated, which has been shown to asymptotically increase to the real ELBO, $\underline{\mathcal{L}}$:
\begin{align}
\underline{\mathcal{L}}=\mathbb{E}_{q(\Xmat)}\mathbb{E}_{q_{\phi}{(\Psimat|\Xmat)}} \mathbb{E}_{ q_{\phi}(\Zmat \given \Psimat, \Xmat)} \log\left( \frac{p(\Xmat,\Zmat)}{q_{\phi}(\Zmat|\Psimat, \Xmat)}\right). 
\label{eqn:ELBO}
\end{align}

An additional regularizer is used to avoid degeneracy:
\begin{equation}\label{eq:lower of SIVI2}
\begin{split}
 B_{K} = \mathbb{E}_{q({\Xmat})} &\mathbb{E}_{{\Psimat},{\Psimat}^{(1)}, \ldots,{\Psimat}^{(K)} \sim q_{{\phiv}}({\Psimat}|\Xmat)}\big[\\
 &\text{KL}(q_{\phi}(\Zmat |\Psimat, \Xmat))||h_K(\Zmat|\Xmat)\big],
 \end{split}
 \end{equation}
where
\begin{equation}\label{eq: hk}
 h_{K}(\Zmat|\Xmat)=\frac{1}{K+1}\left[q_{\phi}\left({\Zmat} | {\Psimat}, \Xmat\right)+\sum_{k=1}^{K} q_{\phi}\left({\Zmat} | {\Psimat}^{(k)}, \Xmat \right)\right]\notag.
\end{equation} 

While the semi-implicit VAE (using the asymptotic lower bound) can capture the correlation and multi-modality among latent dimensions of $\Zmat$, we illustrate in Section~\ref{sec:esihge} that it may compromise the correlation with the input and the corresponding latent representation. We empirically show that we can improve the results by incorporating the mutual information between the input and latent representations during training. This helps the objective function to both preserve the correlation among latent dimensions of the $\Zmat$, as well as to encourage a high level of correlation between the input and its latent representation.

\subsection{Hyperbolic (Poincar\'{e}) Space}
Within the Riemannian geometry framework, hyperbolic spaces are manifolds with constant negative curvature and are of particular interest for embedding hierarchical structures. There are multiple models for an $n$-dimensional hyperbolic space; among these are the hyperboloid $H_{nC}$ (also known as the Lorentz model) (\cite{nickel2018learning}), and the \poincare ball $P_{nC}$ (\cite{nickel2017poincare}). Specifically, in hyperbolic space, distances grow exponentially as a point moves away from the origin, and to describe the shortest path between two distant points, we should pass through a common parent (i.e., the origin) which gives rise to a hierarchical or tree-like structure. In this section, we briefly review the \poincare model for hyperbolic geometry.

A manifold $\mathcal{M}$ is a set of points $\zv$, which are locally similar to a linear space. For Each point $\zv$ on $\mathcal{M}$ we can define its $n$-dimensional \emph{tangent space} $T_{\zv} \mathcal{M}$, which is a first order approximation of the manifold around the point $\zv$. Based on this, we can use an \emph{exponential mapping} to transform a vector in the tangent space at $\boldsymbol v$ to the manifold based on the following formulation:

\begin{equation}\label{eq: exp mapping}
\exp _{\boldsymbol{z}}^{c}(\boldsymbol{v})=\boldsymbol{z} \oplus_{c}\left(\tanh \left(\sqrt{c} \frac{\lambda_{\boldsymbol{z}}^{c}\|\boldsymbol{v}\|}{2}\right) \frac{\boldsymbol{v}}{\sqrt{c}\|\boldsymbol{v}\|}\right), 
\end{equation}

where $c$ is the constant negative curvature, and $\oplus_{c}$ is the \emph{Möbius addition} of two vectors on the manifold and defined as 

\begin{equation}\label{eq: mobius}
\notag
     \boldsymbol{z} \oplus_{c} \boldsymbol{y}=\frac{\left(1+2 c\langle\boldsymbol{z}, \boldsymbol{y}\rangle+c\|\boldsymbol{y}\|^{2}\right) \boldsymbol{z}+\left(1-c\|\boldsymbol{z}\|^{2}\right) \boldsymbol{y}}{1+2 c\langle\boldsymbol{z}, \boldsymbol{y}\rangle+c^{2}\|\boldsymbol{z}\|^{2}\|\boldsymbol{y}\|^{2}}. 
\end{equation}

Conversely, one can map a vector on the manifold to the tangent space using the \emph{logarithmic mapping} defined as  

\begin{equation}\label{eq: log mapping}
    \log _{z}^{c}(\boldsymbol{y})=\frac{2}{\sqrt{c} \lambda_{z}^{c}} \tanh ^{-1}\left(\sqrt{c}\left\|-\boldsymbol{z} \oplus_{c} \boldsymbol{y}\right\|\right) \frac{-\boldsymbol{z} \oplus_{c} \boldsymbol{y}}{\left\|-\boldsymbol{z} \oplus_{c} \boldsymbol{y}\right\|}, 
\end{equation}
where $\lambda_{\zv}^c = \frac{2}{1 - c\|\zv\|^2}$.

Based on the aforementioned mapping, we can map a Gaussian distribution defined in the vector space (tangent space) to the \poincare manifold using the exponential mapping in (\ref{eq: exp mapping}). After passing the Gaussian distribution $\eta \sim \mathcal{N}(\muv, \boldsymbol{\sigma})$ through the exponential mapping we get the \emph{wrapped normal}, defined as
\begin{equation}\label{eq: wrapped normal}
\begin{split}
&\mathcal{N}^{\text{W}}_{\text{p}}(\zv\given\muv, \boldsymbol{\sigmav}) =\\ &\mathcal{N}\left(\lambda_{\boldsymbol{\mu}}^{c} \log _{\boldsymbol{\mu}}(\boldsymbol{z}) | \mathbf{0}, \sigmav\right)\left(\frac{\sqrt{c} d_{p}^{c}(\boldsymbol{\mu}, \boldsymbol{z})}{\sinh \left(\sqrt{c} d_{p}^{c}(\boldsymbol{\mu}, \boldsymbol{z})\right)}\right)^{d-1},
\end{split}
\end{equation}
where $d_{p}^{c}(\boldsymbol{\mu}, \boldsymbol{z})$ is the \poincare distance between $\muv$ and $\zv$, defined as
\begin{equation}\label{eq: poincare distance}
d_{p}^{c}(\boldsymbol{z}, \boldsymbol{\mu})=\frac{1}{\sqrt{c}} \cosh ^{-1}\left(1+2 c \frac{\|\boldsymbol{z}-\boldsymbol{\mu}\|^{2}}{\left(1-c\|\boldsymbol{z}\|^{2}\right)\left(1-c\|\boldsymbol{\mu}\|^{2}\right)}\right).
\end{equation}
	
\section{Enhanced Semi-Implicit Hyperbolic Graph Embedding} \label{sec:esihge}
In this section, we propose a generative model to create a richer latent representation of the graph nodes in hyperbolic space. We model the posterior distribution using the SIVI framework; in addition, we show why semi-implicit VAE's representation can compromise the correlation between input $\Xmat$ and latent code $\Zmat$ and propose a solution to address this problem.
Since we use a semi-implicit wrapped normal representation for our latent model, we start with the semi-implicit VAE objective function given by $\underline{\mathcal{L}}_{K} = \underline{\mathcal{L}}+B_{K}$. Each of these components can be expressed by the following Lemma.

{\bf{Lemma 1:}}\emph{ The semi-implicit VAE objective components can be shown to have the following representation:}
\begin{equation}\label{eq: Lemma1}
\begin{split}
&\underline{\mathcal{L}}_K = \underline{\mathcal{L}} + B_K, \textit{ where:}\\
&\underline{\mathcal{L}} = \mathbb{E}_{ q(\Xmat) q_{\phi}(\Psimat|\Xmat) q(\Zmat|\Psimat,\Xmat)}[\log p(\Xmat| \Zmat)]
   \\ &- I(\Zmat; \Xmat\given\Psimat) - I(\Zmat; \Psimat) - \text{KL}(q(\Zmat)||p(\Zmat)),\\
   &\textit{and } B_{K}\leq I(\Zmat;\Psimat|\Xmat).
\end{split}    
\end{equation}
\emph{Proof is provided in the Supplementary Material.}

In the proof, it has been shown explicitly how each component of the semi-implicit VAE asymptotic lower bound ($\underline{\mathcal{L}}_K = \underline{\mathcal{L}} + B_K$)  is related to the expressions in \eqref{eq: Lemma1}. Next we discuss how the components of the objective function play a role to achieve our goal of a richer latent representation. As shown in Lemma 1, it can be inferred that the maximization of $\underline{\mathcal{L}}_K$ will lead to the maximization of the log-likelihood of reconstruction which is a desirable outcome. Furthermore, the optimization in \eqref{eq: Lemma1} involves the minimization of the mutual information between the observed input and latent representations. As a result, the objective function may reduce the input-latent-representation correlation, specifically when we use very expressive modeling tools such as deep neural networks \citep{rezaabad2020learning}. In addition, it can be observed that maximization of objective will results in minimization of $I(\Zmat;\Psimat)$. However, we note that this term has been compensated by adding the $\lim_{K \rightarrow \infty} B_K$. Indeed, one can show that
\begin{equation}\label{eq: BK is mutual information}
\begin{split}
    & \lim _{K\rightarrow \infty}B_K \leq I(\Zmat;\Psimat)-I(\Xmat;\Zmat;\Psimat), \quad  \text{where} \\ &I(\Xmat;\Zmat;\Psimat) = \int q_{\phi}(\Xmat,\Zmat,\Psimat)\Big[
    \\ &\log  \frac{q_{\phi}(\Xmat,\Zmat) q_{\phi}(\Zmat,\Psimat) q_{\phi}(\Xmat,\Psimat) }{q_{\phi}(\Xmat,\Zmat,\Psimat) q(\Xmat) q_{\phi}(\Zmat) q_{\phi}(\Psimat)} \Big]
    d\Xmat d\Zmat d\Psimat. 
    \end{split}
\end{equation}
Adding  $\lim_{K\rightarrow \infty} B_K$ in the semi-implicit objective function \eqref{eq: Lemma1}, will not address all the concerns which we have discussed earlier. Also, there are certain theoretical and practical drawbacks of adding that term to objective function which we will discuss then in following. 1- Based on \eqref{eq: BK is mutual information}, while the additional term may compensate for the minimization of mutual information between $\Zmat$ and $\Psimat$ ($I(\Zmat;\Psimat)$), it will not address the concern with regard to minimization of $I(\Zmat;\Xmat|\Psimat)$; 2- in addition, it will minimize the mutual information of the tuple $(\Xmat, \Zmat, \Psimat )$; and 3- it is not possible to have the $K$ grow without bound in practice. Specifically, for graph convolution neural networks in which optimization of parameters is conducted on the graph, the semi-implicit VAE lower bound optimization gets prohibitively expensive as $K$ grows. For small $K$s where the computation is practical on large graph data, $B_K$ will not be able to fully compensate for the $I(\Zmat;\Psimat)$ minimization.

Finally, $\text{KL}(q(\Zmat)||p(\Zmat))$ is independent of $\Xmat$ and will not help in modeling the correlation among input and latent representations. In the next section, we discuss how we address the  minimization of the mutual information between the latent representations $\Zmat$ and observed features $\Xmat$.

\subsection{Semi-Implicit Objective Enhancement}
As we discussed, the correlation between (input) and the latent variables may be compromised due to the nature of the semi-implicit VAE objective function. Here we propose a practical and effective enhancement of the semi-implicit VAE by adding a mutual information term to the objective function. It is noteworthy to mention, the added mutual information term should not only address the minimization of mutual information in \ref{eq: Lemma1}, but also, it should be evaluated effectively and efficiently. As mentioned previously, $B_K$ will compensate for the minimization of mutual information between $\Psimat$ and $\Zmat$ when $K$ grows without bound, but the compensation is limited for finite and small values of $K$. Furthermore, as shown in \eqref{eq: Lemma1}, the mutual information between $\Xmat$ and $\Zmat$ is minimized due to the maximization of $\underline{\mathcal{L}}_{K}$. Therefore, to compensate for minimization of $I(\Zmat;\Psimat)$  and $I(\Xmat;\Zmat|\Psimat)$, we add $I((\Xmat, \Psimat);\Zmat)$ to the objective of the semi-implicit VAE. The objective of our \emph{Enhanced Semi-Implicit VAE (ESI-VAE)} is defined as follows:
\begin{equation}\label{eq: Info SIVI}
\begin{split}
    \mathcal{L}_{\text{ESI-VAE}}(\phi) =
    \underline{\mathcal{L}} + 
    B_K + \gamma I(\Xmat, \Psimat;\Zmat),
    \end{split}
\end{equation}
where $\gamma>0$ is a regularization parameter for the mutual information. In practice, $\gamma$ is treated as a hyper-parameter and optimized through cross-validation.
The added term $I(\Xmat, \Psimat;\Zmat)$ can be expanded as  $I(\Zmat;\Psimat) + I(\Xmat;\Zmat|\Psimat)$ which shows how it explicitly compensates for mutual information terms minimized in \eqref{eq: Lemma1}.
The computation of $I(\Xmat, \Psimat;\Zmat)$ imposes a challenge which has been addressed by different works (\cite{esmaeili2019structured, gabrie2018entropy}). The challenge is to approximate the marginal distribution $q_{\phi}(\Zmat)$ which is not computationally tractable. To approximate $q_{\phi}(\Zmat)$ by the samples, one can use Monte-Carlo sampling (\cite{esmaeili2019structured}) over a large amount of data, which could be expensive and cause optimization instability. An alternative solution is to leverage the dual representation of the mutual information (\cite{belghazi2018mine}) which is tractable to compute and stable to optimize. We use $f$-divergence dual estimation to approximate the mutual information lower bound by solving the optimization problem on the space of functions with finite expectations in our latent domain. These methods for estimation offer a stable and computationally efficient technique to approximate the mutual information which can be effectively implemented by drawing samples from the joint and marginals of the posterior.

\subsection{Dual form}
To promote the correlation between input ($\Xmat$) and latent representation ($\Zmat$), the following term is added to the semi-implicit VAE objective function, $I(\Xmat,\Psimat;\Zmat) = \text{KL}(q(\Zmat, \Xmat ,\Psimat) || q(\Zmat) q(\Xmat, \Psimat) )$. We elaborate in the following Lemma:

{\bf{Lemma 2:}} \emph{Substituting the KL-divergence with the variational $f$-divergence, a lower bound for the objective function in (\ref{eq: Info SIVI}) is given by}:
\begin{equation*}\label{eq: dual info sivi}
    \begin{split}
    \mathcal{L}_{\text{ESI-VAE}}(\phi) \geq\\& \max_{T} ~ \underline{\mathcal{L}}_K + \gamma \Big[\mathbb{E}_{q_\phi(\Xmat, \Psimat,\Zmat )}[T(\Xmat, \Psimat, \Zmat)] \\&- \mathbb{E}_{q_\phi(\Xmat,\Psimat)q_\phi( \Zmat)}[f^*(T(\Xmat,\Psimat, \Zmat))]\Big]
    \end{split},
\end{equation*}
\emph{where $f(T) = T\log T$, $f^*$ is the convex conjugate function of $f$, and $T$ represents all possible functions such that their expectations are finite. See Supplementary Material for the proof}.

To evaluate  $\mathbb{E}_{q_{\phi}(\Xmat, \Psimat, \Zmat)}[\ \cdot\ ]$ and  $\mathbb{E}_{q(\Xmat,\Psimat)q_{\phi}(\Zmat)}[\ \cdot\ ]$ in a tractable way, we take the following approach. It is possible to draw samples from $(\Xmat^{(i)},\Psimat^{(i)} \Zmat^{(i)}) \sim q_{\phi}(\Xmat, \Psimat, \Zmat)=q_{\phi}(\Zmat|\Xmat, \Psimat)q_{\phi}(\Psimat|\Xmat)q(\Xmat)$ by first sampling $\epsilonv$ from $q(\epsilonv)$. Having $\epsilonv$ and $\Xmat$ (observed graph), we  can evaluate $\Psimat$ from $\Psimat = g_{\phi}(\Xmat,\epsilonv)$, and then generate $\Zmat$ from $q_{\phi}(\Zmat|\Xmat,\Psimat)$. To create the samples of the marginalized distributions  $q_{\phi}(\Zmat)$ and $q_{\phi}(\Xmat,\Psimat)$, we randomly choose a pair $(\Xmat, \Psimat)^{(j)}$ then sample from $\Zmat\sim q_{\phi}(\Zmat| (\Xmat, \Psimat)^{(i)}),\ i \neq j$. In practice, however, we can effectively get samples from the observed attributes and then permute the representations $\Zmat$. This trick is first used in \cite{arcones1992}, and proved to be sufficiently accurate so long as the training set size is large enough. To be specific, we take correlated tuples of $(\Xmat,\Psimat,\Zmat)$ generated previously, fix the data on the $(\Xmat,\Psimat)$ axis and permute the $\Zmat$ samples to decorrelate $(\Xmat,\Psimat)$ and $\Zmat$ to reflect how the samples of marginalized distributions are generated. It is worth mentioning that the consistency of the estimation we have used relies on the size of the family $T$ and number of samples. Based on the universal approximation theorem for the neural networks (\cite{hornik1989multilayer}), $T$ in Lemma 2 can be closely approximated by neural networks. Moreover, for graph embedding, one need to feed the whole dataset to the networks which leads to near accurate estimation of the two expectations in Lemma 2.

\subsection{Training}
The full training algorithm is described in Algorithm~\ref{Algo1} and the details are discussed in the following subsections.

{\bf{Objective}}: The final objective can be readily extended to graph convolution neural networks defined in \poincare space first by defining $\mathcal{L}^g$ and $B_K^g$:

\begin{align}\label{eq: final objective}
\begin{split}
\mathcal{L}^g=\mathbb{E}_{q(\Xmat, \textbf{A})}\mathbb{E}_{q_{\phi}{(\Psimat|\Xmat, \textbf{A})}} &\mathbb{E}_{ q_{\phi}(\Zmat \given \Psimat, \Xmat, \textbf{A})}[ \\ &\log\left( \frac{p(\textbf{A}, \Xmat,\Zmat)}{q_{\phi}(\Zmat|\Psimat, \Xmat, \textbf{A})}\right)], 
\end{split}
\end{align}

\begin{equation*}
\begin{split}
 B^g_{K} = \mathbb{E}_{q({\Xmat, \textbf{A}})} &\mathbb{E}_{{\Psimat},{\Psimat}^{(1)}, \ldots,{\Psimat}^{(K)} \sim q_{{\phiv}}({\Psimat}|\Xmat, \textbf{A})}\big[\\
 &\text{KL}(q_{\phi}(\Zmat |\Psimat, \Xmat, \textbf{A}))||h^g_K(\Zmat|\Xmat, \textbf{A})\big],
 \end{split}
 \end{equation*}
 where, 
\begin{equation}
\begin{split}
 h^g_{K}(\Zmat|\Xmat, \textbf{A})=\frac{1}{K+1}\big[&q_{\phi}\left({\Zmat}| {\Psimat}, \Xmat, \textbf{A}\right)+\\\sum_{k=1}^{K} &q_{\phi}({\Zmat}| {\Psimat}^{(k)}, \Xmat, \textbf{A} )\big]\notag.
 \end{split}
\end{equation}
Finally, we can formulate our proposed ESI-HGE objective as follows:
\begin{equation}
\begin{split}
    \mathcal{L}_{\text{ESI-HGE}} = &\mathcal{L}^g + B^g_K +\gamma
    \max_{T} \Big[\mathbb{E}_{q_\phi(\Xmat, \Psimat,\Zmat )}[T(\Xmat, \Zmat, \Psimat)] 
    \\ -&\mathbb{E}_{q_\phi(\Xmat,\Psimat)q_\phi( \Zmat)}[f^*(T(\Xmat,\Psimat, \Zmat))]\Big],
\end{split}
\end{equation}
where we seek to estimate $\phi$ as well as $T$ (parameters of the networks) that maximize the objective.  

{\bf{Encoder and decoder}}: Like conventional encoders in GVAEs, the ESI-HGE encoder returns the mean and the variance and then uses the exponential mapping defined in \eqref{eq: exp mapping} to get parameters of $q_{\phi}\left({\Zmat}^{i} | {\Psimat}^{i}, \Xmat, \textbf{A}\right)$, which is defined as a wrapped normal distribution. We use the technique proposed by \cite{poincarevae} to sample from the wrapped normal distribution. For the decoder, our generative model is given by an inner product in the vector space.

{\bf{Mutual information estimation}}: The function $T$ is defined to be a neural network where its inputs are $(\Xmat, \Psimat, \Zmat)$. As $\Psimat$ and $\Zmat$ are in the \poincare space, we first pass them through a \emph{gyroplane} layer to map from the \poincare space to Euclidean space (\cite{ganea2018hyperbolic}).   

\begin{algorithm*} \caption{Enhanced Semi-Implicit Hyperbolic Graph Embedding}
	\label{Algo1}
	\begin{algorithmic}[1]
		\Require $\Xmat, \text{A}$, latent variable dimension $\Zmat_{\text{dim}}$, optimizers: $G_{\phi}$, $G_T$ 
		\State get the adjacency matrix $\textbf{A}$ and nodes attributes $\Xmat$; create train/test/validation 
		\State initialize $\phi$, $T$, G
		\Repeat 
		\State get samples of $\zv \sim q_{\phi}(\Zmat|\Xmat, \Psimat, \textbf{A})$,  $\Psimat \sim q_{\phi}(\Psimat|\Xmat, \textbf{A})$ and create the tuple $\{{(\Xmat, \Psimat, \Zmat})\}_{i=1}^N$, 
		\State {\it permute} the tuple along the $\Zmat$-column  to get $\{{(\Xmat, \Psimat, \hat{\Zmat}})\}_{i=1}^N$
		\State $\phi\leftarrow G\Big[\nabla_\phi \Big(\mathcal{L}^g + B^g_K + \frac{\gamma}{N} \big( \sum_{i=1}^N[T(\Xmat^{(i)}, \Psimat^{(i)}, \Zmat^{(i)}))] 
     -[f^*(T(\Xmat^{(i)},\Psimat^{(i)}, \hat{\Zmat}^{(i)})]\big)\Big)\Big]$
		\State $t\leftarrow G_t\Big[\nabla_t \Big(\frac{1}{N}\sum_{i=1}^{b}T(\Xmat^{(i)}, \Psimat^{(i)}, \Zmat^{(i)}) - \frac{1}{N}\sum_{i=1}^{N}f^*(T(\Xmat^{(i)}, \Psimat^{(i)}, \Zmat^{(i)}))\Big)\Big] $
		\Until convergence
	\end{algorithmic}
\end{algorithm*}

\begin{table*}[h]
	\centering
	\caption{Link prediction performance in networks with node attributes.} 
	\label{tab:link_prediction}
	\resizebox{\textwidth}{!}{\begin{tabular}{l*{7}{c}}
		\toprule[\heavyrulewidth]\toprule[\heavyrulewidth]
		\bf{Method}& \multicolumn{2}{c}{\bf{CORA}}  & \multicolumn{2}{c}{\bf{Citeseer}} &  \multicolumn{2}{c}{\bf{Pubmed}}\\
		&  \multicolumn{1}{c}{AUC} & \multicolumn{1}{c}{AP} &  \multicolumn{1}{c}{AUC} & \multicolumn{1}{c}{AP} &  \multicolumn{1}{c}{AUC} & \multicolumn{1}{c}{AP} \\
		\toprule[\heavyrulewidth]\toprule[\heavyrulewidth]
		DW \citep{perozzi2014deepwalk}& 83.1$\pm$0.01 & 85.0$\pm$0.00 & 80.5$\pm$0.02 & 83.6$\pm$0.01 & 84.4$\pm$0.00 & 84.1$\pm$0.00 \\
		SEAL \citep{zhang2018link} & 90.1$\pm$0.10&83.0$\pm$0.30&83.6$\pm$0.20&77.6$\pm$0.20&96.7$\pm$0.10&90.1$\pm$0.10 \\
		GAE \citep{kipf2016variational} &91.0$\pm$0.02&92.0$\pm$0.03&89.5$\pm$0.04&89.9$\pm$0.05& 96.4$\pm$0.00&96.5$\pm$0.00\\
		GVAE \citep{kipf2016variational}&91.4$\pm$0.01&92.6$\pm$0.01&90.8$\pm$0.02&92.0$\pm$0.02& 94.4$\pm$0.02&94.7$\pm$0.02\\
		G2G \citep{bojchevski2017deep} & 92.1$\pm$0.90&92.6$\pm$0.80&94.5$\pm$0.40&93.2$\pm$0.70& 94.3$\pm$0.30&93.4$\pm$0.50\\
		SIG-VAE\citep{sigvae} & 91.2$\pm$0.05&92.0$\pm$0.10&93.5$\pm$0.02&94.3$\pm$0.10& 96.5$\pm$0.30&96.0$\pm$0.50\\
		\midrule
		\textbf{EVAE} & 91.6$\pm$0.02&92.3$\pm$0.02&94.7$\pm$0.10&94.5$\pm$0.15& -  & -\\
		\textbf{HGE} & 91.8$\pm$0.10&92.7$\pm$0.15&93.1$\pm$0.40&94.2$\pm$0.70& -  & -\\
		\textbf{SI-HGE} & 92.1$\pm$0.01&91.2$\pm$0.10&94.2$\pm$0.10&94.9$\pm$0.10&93.7 $\pm$0.45 & 94.5 $\pm$0.05\\
		\textbf{ESI-HGE} & \textbf{92.4}$\pm$0.20&\textbf{93.2}$\pm$0.02&\textbf{95.9}$\pm$0.10&\textbf{96.2}$\pm$0.2&\textbf{96.6}$\pm$0.15 & \textbf{96.8}$\pm$0.35\\
		\midrule	
	\end{tabular}	}
\end{table*}

\section{Experiments and Results}\label{sec:results}
We evaluated the utility of our proposed graph embedding method by conducting experiments on different analytical tasks: edge link prediction, node classification, latent space interpretability, and mutual information stored in the latent codes.

\subsection{Link Prediction}
We investigate link prediction accuracy of our embedding on three popular citation networks (CORA, Citeseer, and Pubmed) to facilitate comparisons to other state-of-the-art methods. We report the graph data statistics for these networks in the Supplementary Material. We split each dataset into training, validation, and test sets containing, 85\%, 5\%, and 10\% of the total network links, respectively. To avoid performance differences due to random sampling we used identical training/validation/test splits for the comparative methods when possible.

{\bf{Setup}}:
We optimized the network for 1000 epochs, where we used edge link prediction accuracy on the validation set to tune the network hyper-parameters including early stopping. Unless stated differently, in all experiments we set our latent variable dimension to $F=16$, and we use a generative network with one hidden layer of size 32. We compare our method with and without the enhanced SIVI objective (Eq.~\ref{eq: Info SIVI}) to separately evaluate the effect of the additional mutual information term. To aid in fair comparison, we used code from the authors of \cite{kipf2016variational}, \cite{bojchevski2017deep}, and \cite{sigvae} to implement GAE, GVAE, G2G, and SIG-VAE\footnote{We note that implementation details in comparative methods did not always include specific parameter settings necessary for exact replication of their published results. Specifically, we were not able to reproduce the results in \cite{sigvae} using the authors' available code.}. We directly used the results for DW and SEAL published by \cite{perozzi2014deepwalk} and \cite{zhang2018link}, respectively. Due to the prohibitively large size of the Pubmed dataset, we report results directly from the relevant work for this dataset.  We include full implementation details in the Supplementary Material. 

{\bf{Performance}}: Table~\ref{tab:link_prediction} reports the test-set performance measured by average precision (AP) and area under ROC curve (AUC) based on 10 runs with different random initialization for the weights of the neural network. Based on these results, our proposed method outperforms other methods across the citation network, with which we can draw couple of conclusions: 1) The inclusion of a \poincare embedding (SI-HGE in Table~\ref{tab:link_prediction}) shows that the hyperbolic embedding improves over Euclidean SIG-VAE and provides better or comparable performance compared with other methods; and 2) ESI-HGE (obtained by enhancing the objective of SIVI and incorporating the \poincare embedding) significantly outperforms SI-HGE and the other methods across all datasets.

\begin{figure}[h]

	

	\includegraphics[width=1.0\columnwidth]{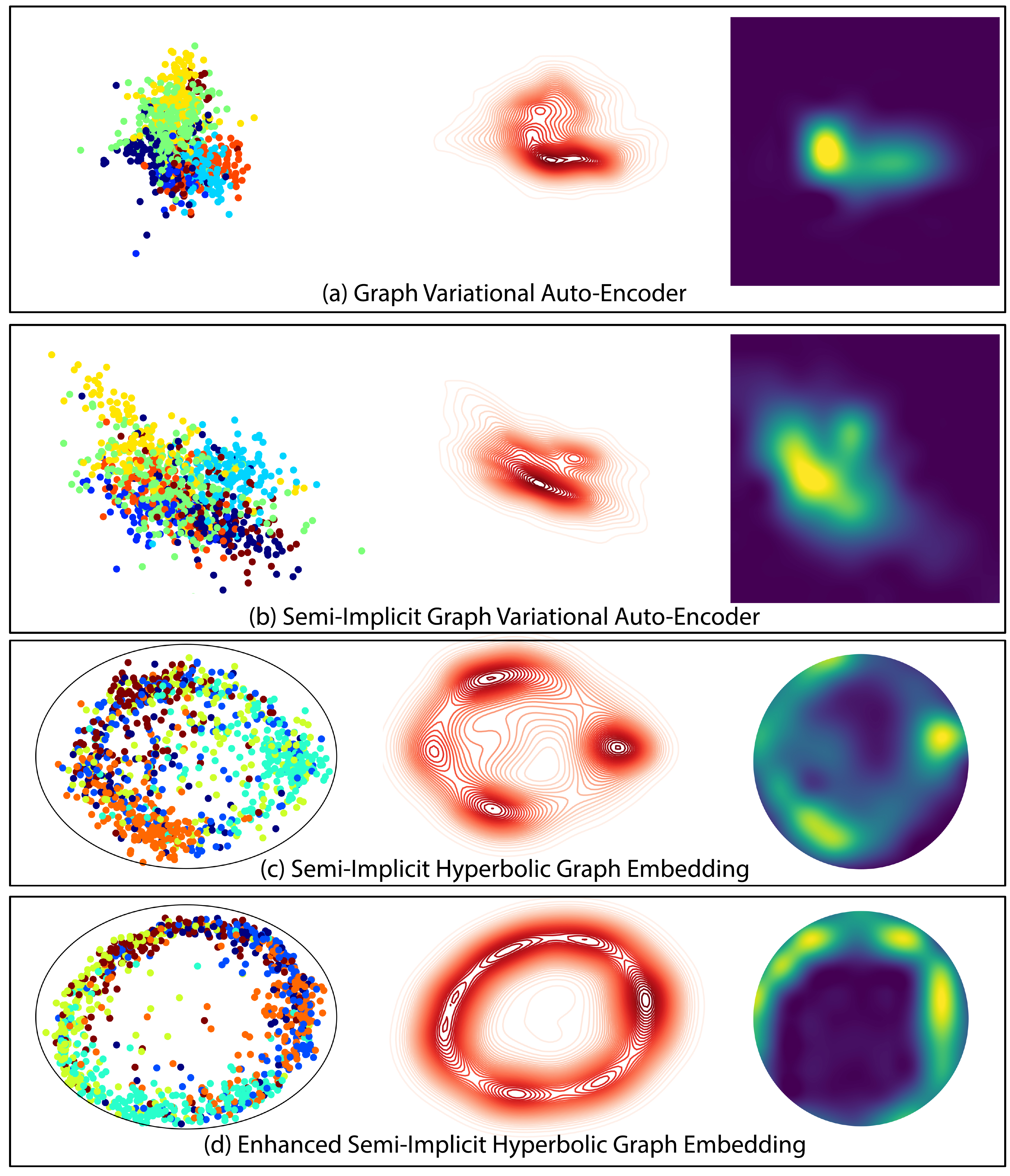}
	\caption{Visualization of Citeseer latent space for (a) GVAE, (b) SIG-VAE, (c) SI-HGE, and (d) ESI-HGE. The GVAE embedding learned a uni-modal distribution concentrated at the origin; SIG-VAE is able to capture additional skewness in the distribution; adding the \poincare embedding provides a more expressive prior with distinct modes, but still shows blending of the different classes (please note that the Citeseer dataset has 6 distinct classes); in  contrast,  ESI-HGE  clearly  enriches  the  posterior representation and is able to recover more expressive distributions with multiple modes distributed across the hyperbolic space
	}
	\label{fig:embedding_visualization}
\end{figure}

{\bf{Visualization}}: To build intuition behind the obtained results, in Figure~\ref{fig:embedding_visualization} we visualize the embedded nodes of GVAE, SIG-VAE, SI-HGE, and ESI-HGE for the Citeseer dataset. The Citeseer citation graph contains five labeled classes (indicated by the colors in the figure). We note that the class labels were not used during training. As the first column of the figure shows, we see that GVAE severely blends the latent codes of the graph for different classes, indicating mode collapsing. SIG-VAE only marginally improves this separation. By endowing the latent space with a hyperbolic geometry, SI-HGE and more significantly ESI-HGE result in more distinct embeddings for each class. In the second and third columns of Figure~\ref{fig:embedding_visualization} we visualize contour plots and heatmaps of the learned posterior distributions. The GVAE embedding recovers a uni-modal distribution concentrated at the origin, while SIG-VAE is able to represent additional skewness in the distribution. Adding the \poincare embedding provides a more expressive prior with distinct modes, but still shows blending of the different classes. In contrast, ESI-HGE clearly enriches the posterior and infers more expressive distributions with multiple modes distributed across the hyperbolic space.

\subsection{Node Classification}
Using the training results from the previous section, we explore classification accuracy on the citation graphs using the node embeddings as input against the labeled classes of each node. We used Cora and Citeseer as described for link prediction, and we omit Pubmed due to limited computational resources. We report the classification accuracy for competing methods from Table 3 in \citep{sigvae}.

{\bf{Setup}}: As previously mentioned, we first trained ESI-HGE in an unsupervised manner without access to class information. After, we train a classifier where its inputs are the learned embeddings and labels are the node classes. To show robustness of the inferred latent codes, we randomly remove 10\% of edges from the graphs to use for test-set performance. Additional details for the classification setup are provided in the Supplementary Material.

{\bf{Performance}}: Classification results are summarized in Table~\ref{tab:classification}. Our proposed method provides a clear improvement in classification accuracy even though it was not explicitly trained for this task.
We note that our setup slightly differs from the methodology in \cite{sigvae}, which explicitly included classification as a semi-supervised loss during VAE training. Thus, we expect our results to further improve under this training procedure.
The result suggests that the learned embeddings in ESI-HGE better capture structure in the graph compared to the other frameworks, which is supported by the richer posterior distribution visualized in Figure~\ref{fig:embedding_visualization}.
\begin{table}[h]
	\centering
	\caption{Classification accuracy on citation graphs.}
	\label{tab:classification}
	\begin{tabular}{l*{4}{c}}
		\toprule[\heavyrulewidth]\toprule[\heavyrulewidth]
		\bf{Method}& \multicolumn{1}{c}{\bf{CORA}}  & \multicolumn{1}{c}{\bf{Citeseer}}  & \multicolumn{1}{c}{\bf{Pubmed}}
 \\
		\toprule[\heavyrulewidth]\toprule[\heavyrulewidth]
		DeepWalk & 67.2 & 43.2 & 65.3 \\
		LP & 68.0 & 45.3 & 63.0 \\
		ICA & 75.1 & 69.1 & 73.9 \\ 
		Planetoid & 75.7 & 64.7 & 77.2\\
		GCN & 81.5 & 70.3 & 79.0 \\
		SIG-VAE & 79.7 & 70.4 & 79.3 \\
		\midrule
		\bf{ESI-HGE} & \bf{84.4}&\bf{72.8} & \bf{82.1}\\
		\midrule	
	\end{tabular}
\end{table}

\subsection{Stored Mutual Information}
{\bf{Setup}}: Another metric that elucidates the usefulness of the learned representation is the information content stored in the embeddings. Using the same citation graphs as previously outlined, here we investigate the mutual information between the observed graph attributes $\Xmat$ and the inferred embeddings $\Zmat$. Similar to node classification, we first trained our model without access to class labels. Next, we used the method proposed by \cite{belghazi2018mine} to estimate the mutual information. More information regarding the setup is provided in the Supplementary Material.

{\bf{Performance}}: The results summarized in Table~\ref{tab:mutualinformation} demonstrate increased mutual information between input and latent spaces due to the enhanced SIVI objective. There is a steady increase in mutual information moving from GVAE to SIG-VAE to SI-HGE to ESI-HGE, suggesting both improved latent representation and successful modeling of the correlation.
\begin{table}[h]
	\centering
	\caption{Mutual information between $(\Xmat, \Psimat), \zv$.} 
	\label{tab:mutualinformation}
	\begin{tabular}{l*{3}{c}}
		\toprule[\heavyrulewidth]\toprule[\heavyrulewidth]
		\bf{Method}& \multicolumn{1}{c}{\bf{CORA}}  & \multicolumn{1}{c}{\bf{Citeseer}} & \multicolumn{1}{c}{\bf{Pubmed}}\\
		\toprule[\heavyrulewidth]\toprule[\heavyrulewidth]
		GVAE & 1.05 & 1.37 & 3.23\\
		SIG-VAE & 2.52 & 2.84 & 4.74\\
		\midrule
		\bf{SI-HGE} & 3.12 & 3.54 & 5.12\\
		\bf{ESI-HGE} & \bf{3.96} &\bf{5.58} & \bf{6.45}\\
		\midrule	
	\end{tabular}
\end{table}

\subsection{Interpretability and Posterior Visualization} 
\begin{figure}[h!]
\centering
\includegraphics[width=0.65\columnwidth]{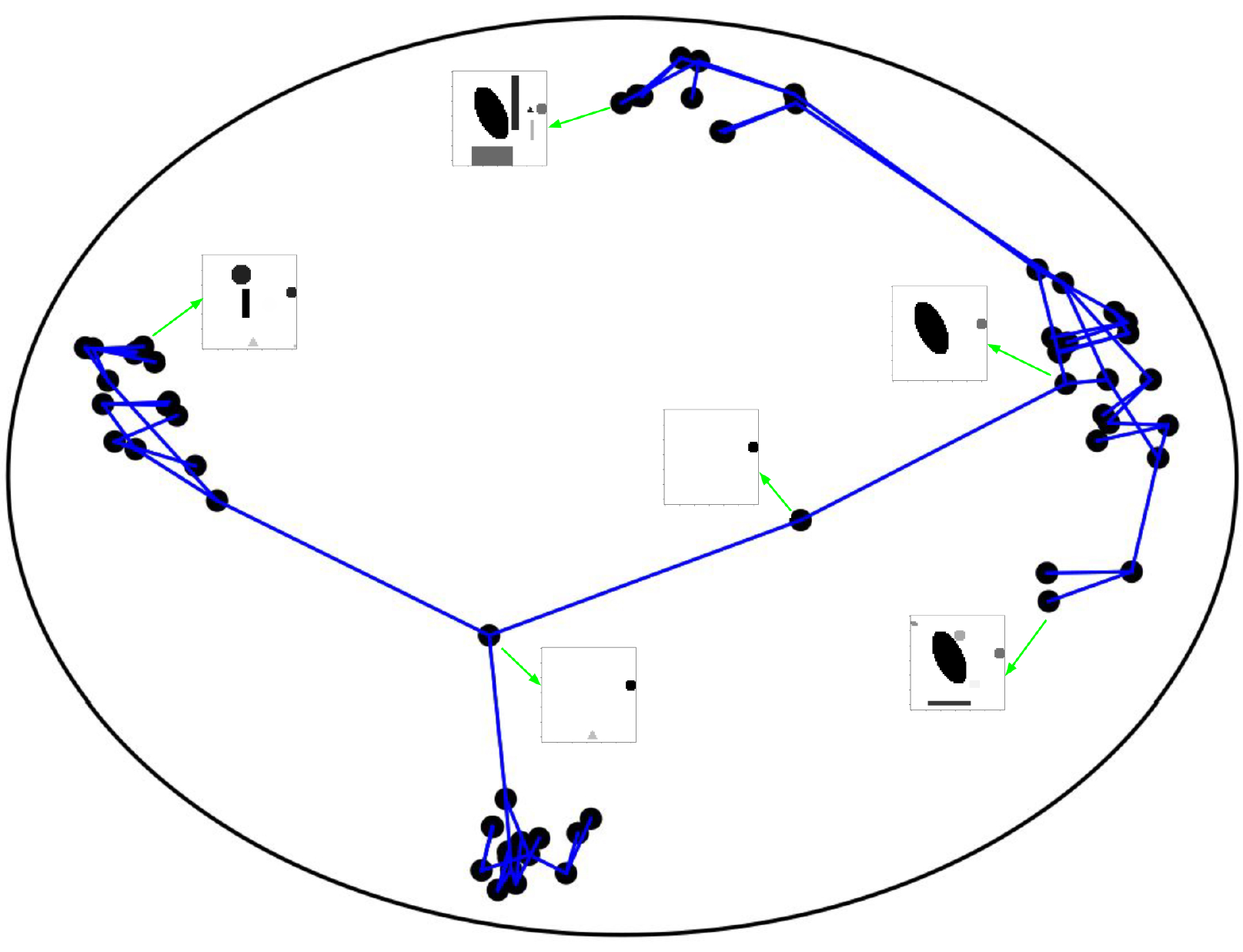}
	\caption{Visualization of the learned latent codes for the synthetic image graph data.}
	\label{fig:image_embedding_paper}
\end{figure}

In the Supplementary Material we outline an additional experiment on a sequence of synthetically generated images forming a graph: we construct a parent image containing a random polygon with random intensity; each successive leaf node is a copy of its parent image with the inclusion of an additional random polygon. We treat the vectorized image pixel intensities as the node features. In Figure~\ref{fig:image_embedding_paper} we visualize the learned latent embedding of ESI-HGE and show a natural hierarchical organization of the nodes that closely matches the true graph structure. This reveals how hyperbolic graph embedding can effectively help to preserve hierarchical representation in latent space.  

\section{Conclusion}\label{sec:conclusion}
We have introduced an enhanced semi-implicit variational auto-encoder for relational graph data with a hyperbolic latent embedding, termed \emph{ESI-HGE}. In addition to representing correlated structures in the latent distribution, our model additionally captures mutual information between the input and the latent space, and thus enhances the learned representations. The theoretical advantages are reflected in our experimental findings, as ESI-HGE outperformed other state-of-the-art methods on edge link prediction and unsupervised node classification, while also increasing interpretability in the latent space.

\section*{Acknowledgments}
This work was supported by  AWS Machine Learning Research Grant, the Office of Naval Research grant N00014-19-1-2590 and Army grant W911NF1910413. 

\bibliography{ref}
\bibliographystyle{abbrvnat}
\setcitestyle{authoryear,open={(},close={)}}

\onecolumn
\begin{center} \textbf{{\Large Hyperbolic Graph Embedding with Enhanced Semi-Implicit Variational Inference}} \end{center}
\appendix
\section{Appendix}
\subsection{Proof of Lemma 1}
To prove, we first start with $\underline{\mathcal{L}}$ from (\ref{eqn:ELBO}):
\begin{align*}
\underline{\mathcal{L}}=&\mathbb{E}_{q(\Xmat)}\mathbb{E}_{q_{\phi}{(\Psimat|\Xmat)}} \mathbb{E}_{ q_{\phi}(\Zmat \given \Psimat, \Xmat)} \log\left( \frac{p(\Xmat,\Zmat)}{q_{\phi}(\Zmat|\Psimat, \Xmat)}\right)\\
=&  \mathbb{E}_{q(\Xmat)}\mathbb{E}_{q_{\phi}{(\Psimat|\Xmat)}} \mathbb{E}_{ q_{\phi}(\Zmat \given \Psimat, \Xmat)} \log\left( \frac{p(\Xmat|\Zmat) p(\Zmat)}{q_{\phi}(\Zmat|\Psimat, \Xmat)}\right)\\
=&  \mathbb{E}_{q(\Xmat)}\mathbb{E}_{q_{\phi}{(\Psimat|\Xmat)}} \mathbb{E}_{ q_{\phi}(\Zmat \given \Psimat, \Xmat)} \log\left( \frac{p(\Xmat|\Zmat) p(\Zmat)q_{\phi}(\Xmat| \Psimat)}{q_{\phi}(\Zmat, \Xmat| \Psimat)}\right)\\
=&  \mathbb{E}_{q(\Xmat)}\mathbb{E}_{q_{\phi}{(\Psimat|\Xmat)}} \mathbb{E}_{ q_{\phi}(\Zmat \given \Psimat, \Xmat)} \log\left( \frac{p(\Xmat|\Zmat) p(\Zmat)q_{\phi}(\Xmat| \Psimat)q_{\phi}(\Zmat| \Psimat)}{q_{\phi}(\Zmat, \Xmat| \Psimat)q_{\phi}(\Zmat| \Psimat)}\right)\\
=&  \mathbb{E}_{q(\Xmat)}\mathbb{E}_{q_{\phi}{(\Psimat|\Xmat)}} \mathbb{E}_{ q_{\phi}(\Zmat \given \Psimat, \Xmat)} \left[ \log p(\Xmat|\Zmat)\right]- \mathbb{E}_{q(\Xmat)}\mathbb{E}_{q_{\phi}{(\Psimat|\Xmat)}} \mathbb{E}_{ q_{\phi}(\Zmat \given \Psimat, \Xmat)}\log\left(\frac{ q_{\phi}(\Zmat \given \Psimat)}{p(\Zmat)}\right) \\
&-\mathbb{E}_{q(\Xmat)}\mathbb{E}_{q_{\phi}{(\Psimat|\Xmat)}} \mathbb{E}_{ q_{\phi}(\Zmat \given \Psimat, \Xmat)}\log\left(\frac{ q_{\phi}(\Zmat, \Xmat \given \Psimat)}{q_{\phi}(\Zmat\given\Psimat)q_{\phi}(\Xmat\given\Psimat)}\right)\\
=&  \mathbb{E}_{q(\Xmat)}\mathbb{E}_{q_{\phi}{(\Psimat|\Xmat)}} \mathbb{E}_{ q_{\phi}(\Zmat \given \Psimat, \Xmat)} \log p(\Xmat|\Zmat)- \mathbb{E}_{q(\Xmat)}\mathbb{E}_{q_{\phi}{(\Psimat|\Xmat)}} \mathbb{E}_{ q_{\phi}(\Zmat \given \Psimat, \Xmat)}\log\left(\frac{ q_{\phi}(\Zmat \given \Psimat)}{p(\Zmat)}\right)\\
&- I(\Zmat; \Xmat \given \Psimat) \\
=&  \mathbb{E}_{q(\Xmat)}\mathbb{E}_{q_{\phi}{(\Psimat|\Xmat)}} \mathbb{E}_{ q_{\phi}(\Zmat \given \Psimat, \Xmat)} \log p(\Xmat|\Zmat)- \mathbb{E}_{q(\Xmat)}\mathbb{E}_{q_{\phi}{(\Psimat|\Xmat)}} \mathbb{E}_{ q_{\phi}(\Zmat \given \Psimat, \Xmat)}\log\left(\frac{ q_{\phi}(\Zmat, \Psimat)}{p(\Zmat)q_{\phi}(\Psimat)}\right)\\
&- I(\Zmat; \Xmat \given \Psimat) \\
=&  \mathbb{E}_{q(\Xmat)}\mathbb{E}_{q_{\phi}{(\Psimat|\Xmat)}} \mathbb{E}_{ q_{\phi}(\Zmat \given \Psimat, \Xmat)} \log p(\Xmat|\Zmat)- \mathbb{E}_{q(\Xmat)}\mathbb{E}_{q_{\phi}{(\Psimat|\Xmat)}} \mathbb{E}_{ q_{\phi}(\Zmat \given \Psimat, \Xmat)}\log\left(\frac{ q_{\phi}(\Zmat, \Psimat)q_{\phi}(\Zmat)}{p(\Zmat)q_{\phi}(\Psimat)q_{\phi}(\Zmat)}\right)\\
&- I(\Zmat; \Xmat \given \Psimat) \\
=&  \mathbb{E}_{q(\Xmat)}\mathbb{E}_{q_{\phi}{(\Psimat|\Xmat)}} \mathbb{E}_{ q_{\phi}(\Zmat \given \Psimat, \Xmat)} \log p(\Xmat|\Zmat)- \mathbb{E}_{q(\Xmat)}\mathbb{E}_{q_{\phi}{(\Psimat|\Xmat)}} \mathbb{E}_{ q_{\phi}(\Zmat \given \Psimat, \Xmat)}\log\left(\frac{ q_{\phi}(\Zmat, \Psimat)}{q_{\phi}(\Psimat)q_{\phi}(\Zmat)}\right)\\
&-\mathbb{E}_{q(\Xmat)}\mathbb{E}_{q_{\phi}{(\Psimat|\Xmat)}} \mathbb{E}_{ q_{\phi}(\Zmat \given \Psimat, \Xmat)} \log \left(\frac{q_{\phi}(\Zmat)}{p(\Zmat)}\right) - I(\Zmat; \Xmat \given \Psimat) \\
=&\mathbb{E}_{q(\Xmat)}\mathbb{E}_{q_{\phi}{(\Psimat|\Xmat)}} \mathbb{E}_{ q_{\phi}(\Zmat \given \Psimat, \Xmat)} \log p(\Xmat|\Zmat) - I(\Zmat; \Psimat) - \text{KL}(q_{\phi}(\Zmat) || p(\Zmat))  - I(\Zmat; \Xmat \given \Psimat).
\end{align*}

Therefore, we have:
\begin{equation*}
 \underline{\mathcal{L}} = \mathbb{E}_{q(\Xmat)}\mathbb{E}_{q_{\phi}{(\Psimat|\Xmat)}} \mathbb{E}_{ q_{\phi}(\Zmat \given \Psimat, \Xmat)} \log p(\Xmat|\Zmat)  - I(\Zmat; \Xmat \given \Psimat) - I(\Zmat; \Psimat) - \text{KL}(q_{\phi}(\Zmat) || p(\Zmat)).
 \end{equation*}

Next, we show that $\lim_{K \rightarrow \infty} B_K \leq I(\Zmat; \Psimat \given \Xmat)$. To prove this, we first note that 
\begin{equation*}\label{eq: appendix lower of SIVI2}
\begin{split}
 B_{k} = \mathbb{E}_{q({\Xmat})} &\mathbb{E}_{{\Psimat},{\Psimat}^{(1)}, \ldots,{\Psimat}^{(K)} \sim q_{{\phiv}}({\Psimat}|\Xmat)}\big[\text{KL}(q_{\phi}(\Zmat |\Psimat, \Xmat))||h_k(\Zmat|\Xmat)\big],
 \end{split}
 \end{equation*}
where
\begin{equation}
 \hv_{k}(\Zmat|\Xmat)=\frac{1}{K+1}q_{\phi}\left({\Zmat} | {\Psimat}, \Xmat\right)+\sum_{k=1}^{K} q_{\phi}({\Zmat} | {\Psimat}^{(k)}, \Xmat )\notag.
\end{equation} 

Therefore, we have:
\begin{align*}
    \lim_{K\rightarrow\infty}B_K &= \lim_{K\rightarrow\infty}  \int q(\Xmat)q_{\phi}(\Psimat|\Xmat)q_{\phi}(\Psimat^{(1)}|\Xmat)\cdots q_{\phi}(\Psimat^{(K)}|\Xmat)q_{\phi}(\Zmat|\Psimat, \Xmat) \log\frac{(K+1)q_{\phi}(\Zmat\given\Psimat, \Xmat)}{q_{\phi}(\Zmat\given\Psimat, \Xmat) + \sum_{k=1}^{K}q_{\phi}(\Zmat\given\Psimat^{(k)},\Xmat)}d\Xmat \cdots\\
    &=  \lim_{K\rightarrow\infty}-\int q(\Xmat)q_{\phi}(\Psimat|\Xmat)q_{\phi}(\Psimat^{(1)}|\Xmat)\cdots q_{\phi}(\Psimat^{(K)}|\Xmat)q_{\phi}(\Zmat|\Psimat, \Xmat) \log\frac{q_{\phi}(\Zmat\given\Psimat, \Xmat) + \sum_{k=1}^{K}q_{\phi}(\Zmat\given\Psimat^{(k)},\Xmat)}{(K+1)q_{\phi}(\Zmat\given\Psimat, \Xmat)}d\Xmat \cdots\\
    & \leq  \lim_{K\rightarrow\infty}-\int q(\Xmat) q_{\phi}(\Psimat|\Xmat)q_{\phi}(\Zmat\given \Psimat, \Xmat) \log \frac{q_{\phi}(\Zmat\given\Psimat, \Xmat) + \sum_{k=1}^K \int q_{\phi}(\Zmat\given\Psimat^{(k)},\Xmat)q_{\phi}(\Psimat^{(k)}\given \Xmat) d\Psimat^{(k)}}{(K+1)q_{\phi}(\Zmat\given \Psimat, \Xmat)}d\Xmat \cdots\\
    & =  \lim_{K\rightarrow\infty}-\int q(\Xmat) q_{\phi}(\Psimat|\Xmat)q_{\phi}(\Zmat\given \Psimat, \Xmat) \log \frac{q_{\phi}(\Zmat\given\Psimat, \Xmat) + \sum_{k=1}^K \int q_{\phi}(\Zmat,\Psimat^{(k)}\given\Xmat)d\Psimat^{(k)}}{(K+1)q(\Zmat\given \Psimat, \Xmat)}d\Xmat d\Psimat d\Zmat
    \\
    & =  \lim_{K\rightarrow\infty}-\int q(\Xmat) q_{\phi}(\Psimat|\Xmat)q_{\phi}(\Zmat\given \Psimat, \Xmat) \log \frac{q_{\phi}(\Zmat\given\Psimat, \Xmat) + K  q_{\phi}(\Zmat\given\Xmat)}{(K+1)q_{\phi}(\Zmat\given \Psimat, \Xmat)}d\Xmat d\Psimat d\Zmat\\
    & =  \lim_{K\rightarrow\infty}-\int q(\Xmat) q_{\phi}(\Psimat|\Xmat)q_{\phi}(\Zmat\given \Psimat, \Xmat) \log \frac{1}{K+1} (1 + K\frac{q_{\phi}(\Zmat\given \Xmat)}{q_{\phi}(\Zmat\given \Psimat, \Xmat)})d\Xmat d\Psimat d\Zmat\\
    & =  -\int q(\Xmat) q_{\phi}(\Psimat|\Xmat)q_{\phi}(\Zmat\given \Psimat, \Xmat) \log  \frac{q_{\phi}(\Zmat\given \Xmat)}{q_{\phi}(\Zmat\given \Psimat, \Xmat)}d\Xmat d\Psimat d\Zmat\\
    & =  \int q(\Xmat) q_{\phi}(\Psimat|\Xmat)q_{\phi}(\Zmat\given \Psimat, \Xmat) \log  \frac{q_{\phi}(\Zmat\given \Psimat, \Xmat)}{q_{\phi}(\Zmat\given \Xmat)}d\Xmat d\Psimat d\Zmat\\
    & =  \int q(\Xmat) q_{\phi}(\Psimat|\Xmat)q_{\phi}(\Zmat\given \Psimat, \Xmat) \log  \frac{q_{\phi}(\Zmat,\Psimat\given  \Xmat)}{q_{\phi}(\Zmat\given \Xmat), q_{\phi}(\Psimat\given \Xmat)}d\Xmat d\Psimat d\Zmat\\
    & = I(\Zmat;\Psimat\given \Xmat)
\end{align*}

The inequality arises from Jensen's inequality applied to the concave function, $\log$. To complete the proof, \cite{sivi} have previously shown that $B_K\leq \lim_{K\rightarrow \infty }B_k$; thus,
\begin{equation*}
    B_K \leq \lim_{K\rightarrow\infty}  B_K \leq I(\Zmat;\Psimat\given \Xmat).
\end{equation*}

Next, we note that
\begin{align*}
    I(\Zmat; \Psimat) - I(\Zmat; \Psimat\given\Xmat) &= \int q_{\phi}(\psi, \Zmat) \log \frac{q_{\phi}(\Psimat, \Zmat)}{q_{\phi}(\Psimat)q_{\phi}(\Zmat)} d\Zmat d\Psimat  - \int q_{\phi}(\Xmat, \Psimat, \Zmat) \log \frac{q_{\phi}(\Zmat, \Psimat\given \Xmat)}{q_{\phi}(\Zmat\given \Xmat)q_{\phi}(\Psimat\given \Xmat)}d\Xmat d\Psimat d\Zmat\\
    &= \int q_{\phi}(\Xmat, \psi, \Zmat) \log \frac{q_{\phi}(\Psimat, \Zmat)}{q_{\phi}(\Psimat)q_{\phi}(\Zmat)}d\Zmat d\Psimat d\Xmat +  \int q_{\phi}(\Xmat, \Psimat, \Zmat) \log \frac{q_{\phi}(\Zmat\given \Xmat)q_{\phi}(\Psimat\given \Xmat)}{q_{\phi}(\Zmat, \Psimat\given \Xmat)}d\Xmat d\Psimat d\Zmat\\
    &= \int q_{\phi}(\Xmat, \psi, \Zmat) \log \frac{q_{\phi}(\Psimat, \Zmat)}{q_{\phi}(\Psimat)q_{\phi}(\Zmat)} \times \frac{q_{\phi}(\Zmat\given \Xmat)q_{\phi}(\Psimat\given \Xmat)}{q_{\phi}(\Zmat, \Psimat\given \Xmat)}d\Xmat d\Psimat d\Zmat\\
    &= \int q_{\phi}(\Xmat, \psi, \Zmat) \log \frac{q_{\phi}(\Psimat, \Zmat)}{q_{\phi}(\Psimat)q_{\phi}(\Zmat)} \times \frac{q_{\phi}(\Zmat, \Xmat)q_{\phi}(\Psimat, \Xmat)q(\Xmat)}{q_{\phi}(\Xmat, \Psimat, \Zmat)q(\Xmat)q(\Xmat)}d\Xmat d\Psimat d\Zmat\\
    &= \int q_{\phi}(\Xmat, \psi, \Zmat) \log \frac{q_{\phi}(\Psimat, \Zmat)q_{\phi}(\Zmat, \Xmat)q_{\phi}(\Psimat, \Xmat)}{q(\Xmat)q_{\phi}(\Psimat)q_{\phi}(\Zmat)q_{\phi}(\Xmat, \Psimat, \Zmat)} d\Xmat d\Psimat d\Zmat\\
    &= I(\Xmat;\Psimat;\Zmat).
\end{align*}
Therefore,
\begin{align*}
    &I(\Zmat; \Psimat) - I(\Zmat; \Psimat\given\Xmat) = I(\Xmat; \Psimat;\Zmat)\\
    \implies& \lim_{K\rightarrow\infty} B_K \leq I(\Zmat;\Psimat) - I(\Xmat;\Psimat;\Zmat).
\end{align*}

\subsection{Proof of Lemma 2:}
First, we note that
\begin{equation*}
\begin{split}
    \mathcal{L}_{\text{ESI-VAE}}(\phi) &=
    \underline{\mathcal{L}} + 
    B_K + \gamma I(\Xmat, \Psimat;\Zmat)\\
    \mathcal{L}_{\text{ESI-VAE}}(\phi) &= \underline{\mathcal{L}}_K + \gamma I(\Xmat, \Psimat;\Zmat)\\
    &=
    \underline{\mathcal{L}}_K + \gamma \text{KL}(q_{\phi}(\Xmat, \Psimat, \Zmat) || q_{\phi}(\Xmat, \Psimat)q_{\phi}(\Zmat)).
\end{split}
\end{equation*}
By replacing the KL-divergence with $f$-divergence (by assuming  $f(T)=T\log T$) \citep{nowozin2016f}, we have:
\begin{equation*}
\begin{split}
    \mathcal{L}_{\text{ESI-VAE}}(\phi) &=\underline{\mathcal{L}}_K + \gamma D_f(q_{\phi}(\Xmat, \Psimat, \Zmat) || q_{\phi}(\Xmat, \Psimat)q_{\phi}(\Zmat))\\
    &=\underline{\mathcal{L}}_K + \gamma \mathbb{E}_{q_{\phi}(\Xmat, \Psimat)q_{\phi}(\Zmat)}\left[f(\frac{q_{\phi}(\Xmat, \Psimat, \Zmat)}{q_{\phi}(\Xmat, \Psimat)q_{\phi}(\Zmat)})\right]\\
     &=\underline{\mathcal{L}}_K + \gamma \int {q_{\phi}(\Xmat, \Psimat)q_{\phi}(\Zmat)}f(\frac{q_{\phi}(\Xmat, \Psimat, \Zmat)}{q_{\phi}(\Xmat, \Psimat)q_{\phi}(\Zmat)}) d\Xmat d\Psimat d\Zmat.
    \end{split}
\end{equation*}
Next, we replace the $f$-function with its convex conjugate function, $f^*$:
\begin{equation*}
\begin{split}
    \mathcal{L}_{\text{ESI-VAE}}(\phi)   &=\underline{\mathcal{L}}_K + \gamma \int q_{\phi}(\Xmat, \Psimat)q_{\phi}(\Zmat) ~~ \max_{T} \{ ~ T(\Xmat, \Psimat, \Zmat) \frac{q_{\phi}(\Xmat, \Psimat, \Zmat)}{q_{\phi}(\Xmat, \Psimat)q_{\phi}(\Zmat)}- f^*(T(\Xmat, \Psimat, \Zmat))\} d\Xmat d\Psimat d\Zmat\\
    & \geq \underline{\mathcal{L}}_K + \gamma \max_T \{\int q_{\phi}(\Xmat, \Psimat, \Zmat) T(\Xmat, \Psimat, \Zmat)  - q_{\phi}(\Xmat, \Psimat)q_{\phi}(\Zmat) f^*(T(\Xmat, \Psimat, \Zmat)) d\Xmat d\Psimat d\Zmat\}
    \\
    & = \underline{\mathcal{L}}_K + \gamma \max_T \{\mathbb{E}_{q_{\phi}(\Xmat, \Psimat, \Zmat)}  [T(\Xmat, \Psimat, \Zmat)]  - \mathbb{E}_{q_{\phi}(\Xmat, \Psimat)q_{\phi}(\Zmat)}[ f^*(T(\Xmat, \Psimat, \Zmat))] \}.
    \end{split}
\end{equation*}
where the inequality arises from Jensen's inequality applied to the convex function, $\max$.

\section{Datasets}
In this paper we use three well-known citation networks. Their statistics are summarized in Table \ref{tab:datasets}. 
\begin{table}[h]
	\centering
	\caption{Citation network graph statistics}
	\label{tab:datasets}
	\begin{tabular}{l|c c c c}
		\toprule[\heavyrulewidth]\toprule[\heavyrulewidth]
		Dataset& Nodes  & Edges & Features & Classes \\
		\toprule[\heavyrulewidth]\toprule[\heavyrulewidth]
		\bf{Cora} & 2,708 & 5,429 & 1,433 & 7\\
		\bf{Citseer} & 3,327 & 4,732 & 3,703  & 6\\
		\bf{Pubmed} & 19,717 & 44,338 &  500  & 3\\
		\hline
	\end{tabular}
\end{table}
The class labels are used to visualize the results in Figure~\ref{fig:embedding_visualization} and to train the classifiers in Table~\ref{tab:classification}. The labels were not used while training the auto-encoders.

\section{Implementation Details}
Reference code to reproduce our results is available at \url{https://github.com/utcsilab/ESI_HGE}.

We use one graph convolutional network of size 32 and one graph convolutional network of size 16 for modeling the mean and covariance parameters of the wrapped normal distribution in \poincare space, $\muv$ and $\boldsymbol{\Sigma}$. We follow the approach of \cite{sivi} and inject Bernoulli noise (to model $\epsilonv$) of dimension 32 to the layer that infers $\muv$. We performed a grid search over the $J$ and $K$ parameters for SIG-VAE for the Cora and Citeseer datasets. For ESI-HGE, we performed a separate grid search over $J$, $K$, and $c$, followed by an additional grid search over $c$ and $\gamma$. For all experiments, we used the Adam optimizer with learning rate $10^{-5}$, $\beta_1 = 0.9$, and $\beta_2 = 0.999$. For ESI-HGE applied to Cora, we set $J=3$, $K=18$, $c=3.1$, and $\gamma=1$. For Citeseer, we set $J=3$, $K=11$, $c=1.6$, and $\gamma=1$. We did not perform grid search for Pubmed due to limited computational resources; instead, we set $J=3$, $K=18$, and used $c=3.1$ and $\gamma=1$ to match the other Cora hyperparameters. For Citeseer and Cora, we repeated each experiment 10 times with different random initialization of the weights and report mean and standard deviation results. Due to limited computational resources, for Pubmed we performed only one experiment. For all experiments we used a latent space of dimension 16.

To estimate the mutual information $I(\Xmat, \Psimat; \Zmat)$ we used a network described by Table \ref{tap: network T}.
 \begin{table}
	\caption{The neural network to model $T$ and estimate the mutual information} 
	\label{tap: network T}
	\centering 
	\begin{tabular}{l} 
		\toprule[\heavyrulewidth]\toprule[\heavyrulewidth]
		\textbf{Architecture}\\
		\midrule
		Input: Concatenation of $\Zmat$, $\Psimat$, $\Xmat$ \\
		FC $1000$ ReLU\\
		FC $400$ ReLU\\
		FC $100$ ReLU\\
		FC $1$ Linear\\
		\midrule
		\bottomrule[\heavyrulewidth] 
	\end{tabular}
\end{table}

\section{Synthetic Dataset}
\begin{figure}[h]
\centering
	\includegraphics[width=.9\columnwidth]{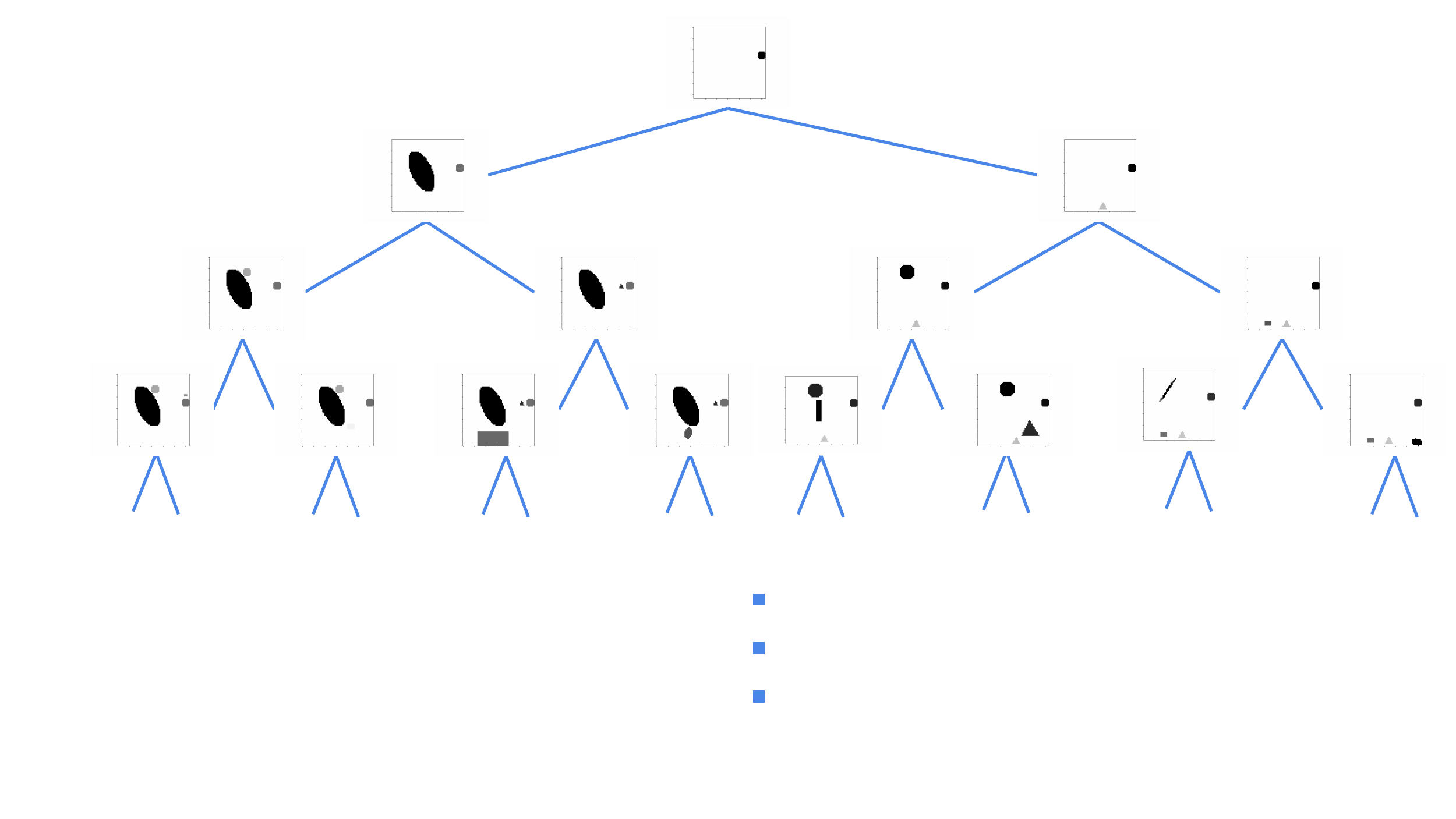}
	\caption{Visualization of our synthetically generated tree containing a natural hierarchy. Each node is represented as an image whose features are the vectorized pixel intensities. Child nodes are images containing the same pixel intensities of its parent node, with an additional non-overlapping random geometric shape of random intensity.}
	\label{fig:image_graph}
\end{figure}

We assessed our proposed ESI-HGE method on synthetically generated image data with a hierarchical structure. We build a balanced binary tree where each node is a grayscale $64\times 64$ image. The parent node image contains a single random geometric shape with random intensity. Each subsequent node has two children. Each child is a copy of its parent image, with an additional non-overlapping random geometric shape with random intensity inserted in the field of view. The graph consisted of 64 total images. Figure~\ref{fig:image_graph} depicts the tree structure for the first 15 nodes with the corresponding images. We vectorize the pixel values of each image to form a length-$4096$ feature vector. We use the same network architecture for the synthetic data as was used for the citation network data in Section~\ref{sec:results}. For network hyperparameters we set $J=3$, $K=18$, $c=1$, and $\gamma=1$. We set the dimension of the latent space to $2$ to aid in visualization.

\clearpage

\subsection*{Results on Synthetic Data}
\begin{figure}[h!]
\centering
\includegraphics[width=0.6\columnwidth]{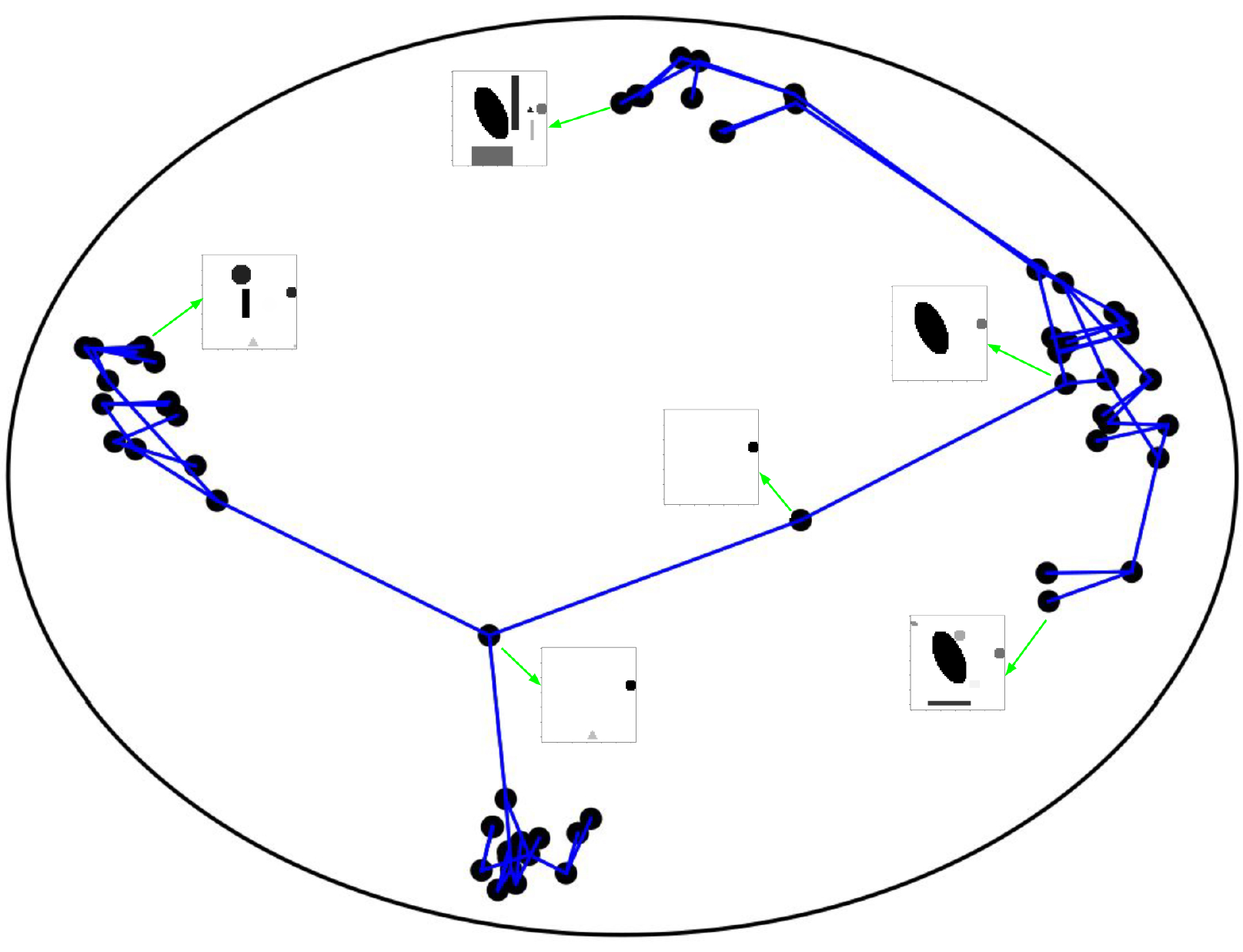}
	\caption{Visualization of the learned latent codes for the synthetic image graph data. The hierarchical structure is naturally preserved.}
	\label{fig:image_embedding}
\end{figure}

\begin{figure}[h!]
\centering
	\includegraphics[width=0.6\columnwidth]{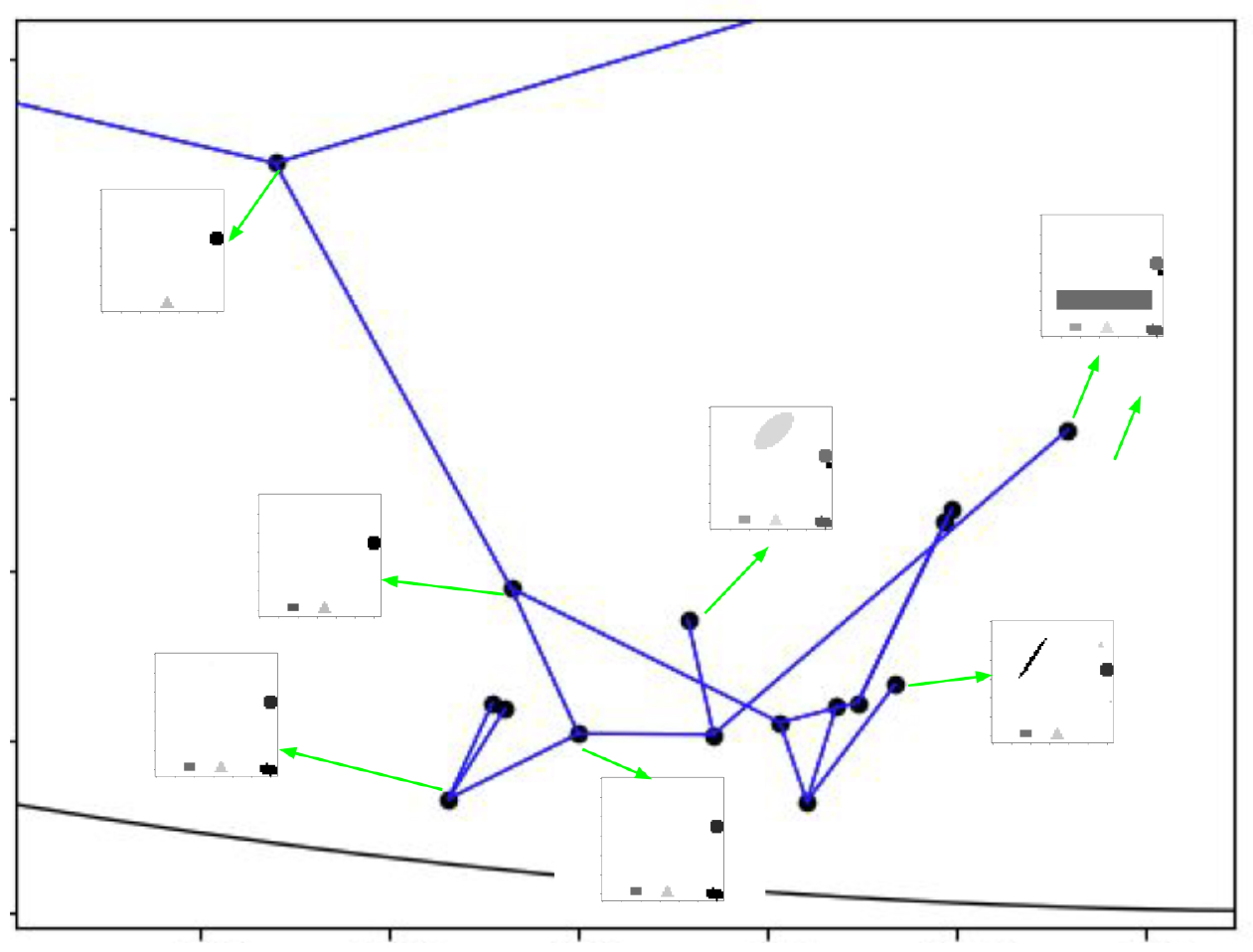}
	\caption{Zoom-in of Figure~\ref{fig:image_embedding} depicting the learned latent codes for the synthetic image graph data.}
	\label{fig:image_embedding_zoom}
\end{figure}

In Figure~\ref{fig:image_embedding} we visualize the latent embedding of the graph with true edges and images overlayed. As the figure shows, a natural hierarchical organization arises in the latent space, where nodes in different branches of the tree occupy different spaces in the embedding.
Figure ~\ref{fig:image_embedding_zoom} shows a zoomed-inset of latent space for one particular branch of the tree, where similar images map to similar embeddings. We note that some child nodes are nearly identical to their parent node except at a few pixels due to the addition of a small random shape only affecting a small number of pixels.

  
  
  

\end{document}


%

%

\onecolumn
\aistatstitle{Instructions for Paper Submissions to AISTATS 2021: \\
Supplementary Materials}

\section{FORMATTING INSTRUCTIONS}

To prepare a supplementary pdf file, we ask the authors to use \texttt{aistats2021.sty} as a style file and to follow the same formatting instructions as in the main paper.
The only difference is that the supplementary material must be in a \emph{single-column} format.
You can use \texttt{supplement.tex} in our starter pack as a starting point, or append the supplementary content to the main paper and split the final PDF into two separate files.

Note that reviewers are under no obligation to examine your supplementary material.

\section{MISSING PROOFS}

The supplementary materials may contain detailed proofs of the results that are missing in the main paper.

\subsection{Proof of Lemma 3}

\textit{In this section, we present the detailed proof of Lemma 3 and then [ ... ]}

\section{ADDITIONAL EXPERIMENTS}

If you have additional experimental results, you may include them in the supplementary materials.

\subsection{The Effect of Regularization Parameter}

\textit{Our algorithm depends on the regularization parameter $\lambda$. Figure 1 below illustrates the effect of this parameter on the performance of our algorithm. As we can see, [ ... ]}

\vfill